\documentclass{Interspeech2024}
\usepackage{multirow}
\usepackage{amsmath}
\usepackage[noend]{algpseudocode}
\MakeRobust{\Call}
\usepackage{algorithm}

\interspeechcameraready

\title{Speech ReaLLM – Real-time Streaming Speech Recognition\\
with Multimodal LLMs by Teaching the Flow of Time}

\name{Frank}{Seide}
\name{Morrie}{Doulaty}
\name{Yangyang}{Shi}
\name{Yashesh}{Gaur}
\name{Junteng}{Jia}
\name{Chunyang}{Wu}

\address{
Meta, USA
\email{\{seide,mdoulaty,yyshi,yashgaur,junteng,chunyang\}@meta.com}
}

\begin{document}

\maketitle

\begin{abstract}
We introduce {\em Speech ReaLLM}, a new ASR architecture that marries ``decoder-only'' ASR with the RNN-T to make multi-modal LLM architectures capable of real-time streaming. This is the first ``decoder-only'' ASR architecture designed to handle continuous audio without explicit end-pointing.
Speech ReaLLM is a special case of the more general {\em ReaLLM} (``real-time LLM'') approach, also introduced here for the first time. The idea is inspired by RNN-T: Instead of generating a response only at the end of a user prompt, generate {\em after every input token} received {\em in real time} (it is often empty).
On Librispeech ``test,'' an 80M Speech ReaLLM achieves WERs of 3.0\% and 7.4\% in real time (without an external LM or auxiliary loss). This is only slightly above a 3x larger Attention-Encoder-Decoder baseline. We also show that this way, an LLM architecture can learn to represent and reproduce the {\em flow of time}; and that a pre-trained 7B LLM can be fine-tuned to do reasonably well on this task.






\end{abstract}

\section{Introduction}

It is the year 2024.
Despite unprecedented progress in AI with Large Language Models,
Artificial General Intelligence remains elusive.
Experts highlight a fundamental limitation of LLMs: they are ``not coupled in real time with the world'' \cite{bach2024}.

This paper introduces a new way of using multi-modal LLM architectures for processing input in a {\em real-time streaming fashion}---not by changing the model architecture itself, but by extending {\em how the model is used and trained}.
We refer to this as {\em the ReaLLM} for ``real-time LLM.''

We test the ReaLLM approach on the task of automatic speech recognition (ASR). We call that variant the {\em Speech Rea\-LLM}. On Librispeech, we find, for both a ``toy''-sized LLMs trained from scratch and a fine-tuned pre-trained LLM, that Speech ReaLLM is a viable architecture for real-time, streaming ASR---the first ``decoder-only'' ASR architecture designed for continuous audio input without explicit end-pointing.
Note that we use the term LLM loosely to mean {\em generative ``decoder-only'' architectures} as used by popular LLM-based chatbots, but without implying pre-training or billions of parameters.

\subsection{The problem with LLMs}

Today's LLM-based chat bots like ChatGPT, Claude, or Meta AI all operate turn-by-turn.
A user types a prompt and hits enter; or speaks a prompt and pauses long enough for endpointing to trigger.
Only then will the LLM encode (``pre-fill'') the prompt, and invoke the LLM response-generation routine which will generate tokens until a special end-of-sentence (EOS) token has been predicted.
These LLMs are {\em reactive}.

Enabling LLMs to operate in real time---to be {\em pro-active}---opens up applications from small time scales such as real-time transcription or acoustic-event detection; via sentence-scale speech scenarios like natural dialogs; to long-span processes like monitoring sensors via a spoken standing prompt.%
\noindent\begin{figure*}[t]
\footnotesize
\begin{tabular}{p{0.33\textwidth}p{0.33\textwidth}p{0.33\textwidth}}
\multicolumn{3}{c}{\hspace*{-1cm}\includegraphics[width=1.0\textwidth,trim={0 7.2cm 0 7.7cm},clip]{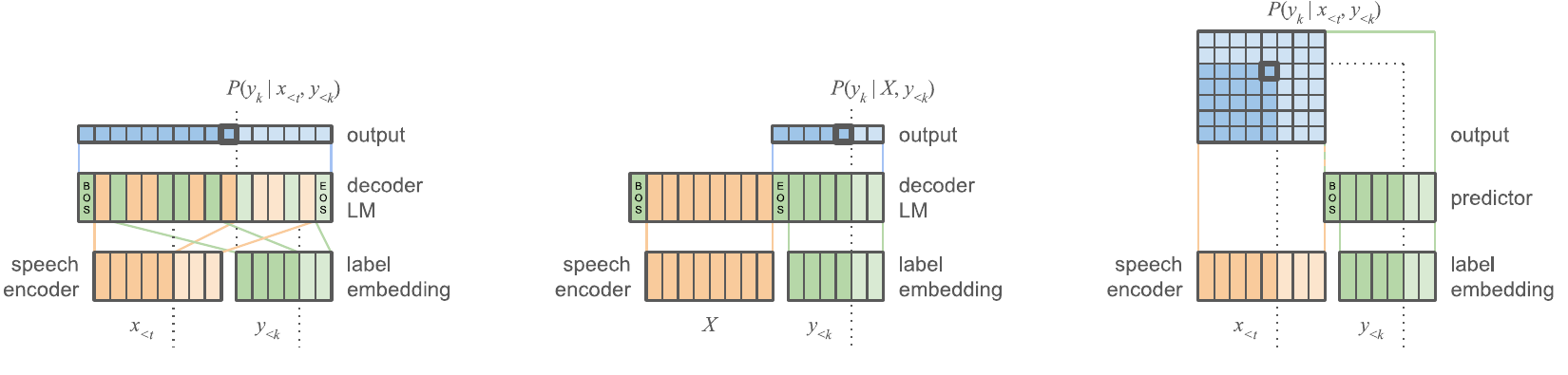}} \\
\centering{\hspace*{-1cm}(a)} Speech ReaLLM (this paper) & \centering{\hspace*{-1cm}(b) Speech LLM \cite{wu2023decoder,fathullah2023prompting}} & \centering{\hspace*{-1cm}(c) RNN-T \cite{graves2012} with joint-network \cite{rnntwithjoiner}}
\end{tabular}
\caption{Comparison of how the three system architectures process and combine encoded speech chunks and decoded labels.}
\vspace*{-4ex}
\label{fig:arch}
\end{figure*}
\subsection{The ReaLLM (Real-time LLM)}

We propose to use and train LLMs differently: Instead of invoking generation at the end of a prompt, invoke generation {\em after every single input token received in real time} to immediately generate a ``response''---although responses would be the empty string most of the time, and non-empty only when appropriate. 

One can see how this can implement real-time ASR if a ``token'' is a chunk of newly received speech while the user is still speaking; and the ``response'' is any new word(s) heard (or empty).\footnote{Although we are concerned with ASR,
ReaLLM is not limited to speech. Input tokens could be A/V feeds from a home security camera (``Let me know if you hear the sound of smashing glass in my home''), health sensors, etc.}
Readers familiar with Recurrent Neural-Network Transducers, or RNN-Ts \cite{graves2012}, will recognize this as a quite natural generalization of the RNN-T's BLANK mechanism.

\subsection{The Speech ReaLLM}

We validate ReaLLM on ASR, for which well-known data sets and evaluation metrics exist.
{\em Speech Rea\-LLM} shall mean applying the Rea\-LLM idea to streaming ASR.
%
Unlike non-streaming decoder-only ASR models \cite{wu2023decoder,fathullah2023prompting} or Whisper \cite{Radford2023-cv}, Speech Rea\-LLM does not see the entire speech utterance from the start. Rather,
speech is revealed chunk by chunk each time the LLM's generation loop terminates,
realizing streaming.

\subsection{Related Work}

To our best knowledge, this combination of LLM/decoder-only based ASR with streaming has not been done before.
The closest are the RNN-T \cite{graves2012} and streaming variants of
Attention-Encoder-Decoder (AED) models like \cite{communication2023seamlessm4t}
that use an explicit READ/WRITE model to alternate between generating output (WRITE) and receiving new input (READ).
Unlike Speech Rea\-LLM, both RNN-T and streaming AED require complex loss functions or learning algorithms like Monotonic Multihead Attention \cite{mma}, Monotonic Chunkwise Attention \cite{mocha}, Monotonic attention with Infinite Lookback \cite{milk}, or EMMA \cite{emma}.

Besides these, literature about ``real time LLMs'' tends to refer to unrelated topics of either re-active use of LLMs with real-time databases, or efficient LLM evaluation.

In the following, we introduce the Speech Rea\-LLM streaming model in detail and how to train it. We then present results comparing Speech ReaLLM on Librispeech to its two ancestors, Speech LLM and RNN-T, and a fine-tuned 7B LLM.

\def\b{\hspace*{-1ex}}

\section{``Decoder-Only'' Streaming ASR}

Speech ReaLLM can be colloquially described as ``Speech LLM and RNN-T having a baby,''
where with Speech LLM, we refer to non-streaming multi-modal ``decoder-only''\footnote{%
``Decoder-only'' is in quotes because this term has come to denote models
without cross-attention into an encoder.
Like most MM-LLMs, Speech ReaLLM surely has an encoder, just not via cross-attention.
} ASR architectures \cite{wu2023decoder,fathullah2023prompting}.
{\bf From Speech LLMs}, Speech ReaLLM inherits the decoder, which centers around a stack of Llama-2 type Transformer decoder layers \cite{touvron2023llama2}, and a multi-modal encoder that encodes speech input into a sequence of embedding vectors used in place of/mixed with text embeddings.\footnote{If we use a pre-trained LLM, the encoder would learn to ``fool'' the LLM into treating the speech embeddings like text tokens.}
 For real-time ASR, a streaming encoder is required, such as an Emformer \cite{shi2021emformer}, Conformer \cite{Gulati2020-vv}, or Streaming Conformer \cite{Shi2022-iu}.
{\bf From the RNN-T}, Speech ReaLLM inherits the {\em BLANK token}, which the RNN-T emits to indicate that no more tokens can be generated without first receiving more speech input.

Speech ReaLLM's operation is best illustrated by its greedy inference algorithm:\footnote{Extension to beam search is left as an exercise to the reader.}

\vspace*{-1.5ex}
\begin{algorithm}[h]
\caption{ReaLLM Greedy Inference}
\label{alg:greedy}
\begin{algorithmic}[1]
\State $h\gets$ [\Call{embed token}{BOS}]
\While{$e\gets$\Call{await \& embed next real-time input}{}} \label{code:receive}
\State $h.$\Call{add}{$e$}
\While{$w\gets$\Call{predict token}{h}, $w\ne$ BLANK} \label{code:inner}
  \State $h.$\Call{add}{\Call{embed token}{w}} \label{code:endinner}
\EndWhile
\EndWhile
\State $h.$\Call{add}{\Call{embed token}{EOS}} \label{code:last}
\While{$w\gets$\Call{predict token}{h}, $w\ne$ EOS}
  \State $h.$\Call{add}{\Call{embed token}{w}}
\EndWhile
\end{algorithmic}
\end{algorithm}
\vspace*{-1.5ex}

The algorithm processes speech input one embedding vector $e$ at a time, where
in a real-time setting, line \ref{code:receive} would block until sufficient additional audio
data has been received to produce the next embedding vector. In our case, an input embedding is generated every 240 ms of audio.

Each time a speech embedding has been received, it is added to the LLM history $h$.
Unlike traditional (non-real-time) LLM decoding, however, we now {\em immediately} perform text generation (line \ref{code:inner} ff.)
until a special BLANK token has been predicted.
When no new words were received, the model would immediately predict BLANK, ending the loop right away.%
\footnote{If we delete the inner loop (line \ref{code:inner}), we get non-streaming inference.}

Note that the first \ref{code:endinner} lines are sufficient for streaming transcription of a continuous audio stream in real time; but to decode audio files, we need to allow the model to emit additional trailing tokens at the end (line \ref{code:last} ff.). The end of speech input is communicated to the decoder as an EOS embedding.


\section{Training The Speech ReaLLM}

The training objective for Speech ReaLLM is not immediately obvious due to the need for time alignment.
An efficient algorithm like the RNN-T loss, which marginalizes over all possible alignments via forward-backward, does not exist because the decoder-only structure cannot be factorized accordingly.

\def\bkr{\rule{1.7mm}{.5pt}}
\def\bk{\rule{1.7mm}{.5pt} }
\def\b{\hspace*{-1ex}}

We approximate such loss by using fixed alignments generated by an external CTC \cite{Graves2006-bq} model, the  ``alignment-teacher.''
Initially we considered to delay the label emissions by, say, half a second, to give the model an opportunity to see some limited future context.
However it turned out to be more effective to provide future context acoustically in the streaming encoder.

Consider this time-aligned example training utterance:
\begin{center}
\def\m#1{\texttt{\footnotesize #1}}
\footnotesize
\begin{tabular}{|r||c|c|c|c|c|c|c|c|}
\hline
Word     &\m{and}&\b\texttt{hand}\b&\m{it}&\b\m{over}\b&\m{to}&\m{you}&end \\ \hline
\b$t_\mathrm{start}$ [ms]&\b 140\b&\b 460\b& 740 &    900\b&\b 1180\b&\b 1380\b&\b 2180\b \\ \hline
$t_\mathrm{end}$ [ms]&\b 380\b&\b 740\b& 860 &\b 1180\b
&\b 1380\b&\b 1700\b& \\ \hline
\end{tabular}
\end{center}
To form the training target sequence, we convert this into a label sequence where a BLANK symbol (denoted as \bkr) stands for an embedding vector that represents 240 ms of speech:
\begin{center}
\noindent\texttt{\footnotesize
\bk \bk and \bk \bk hand it \bk  over \bk   to \bk \bk you \bk \bk EOS
}
\end{center}
We derive the embedding sequence at the input of the LLM decoder by embedding each word label, while for each BLANK, we substitute the speech embedding for the corresponding time:
\def\sp{\hspace*{0.55ex}}
\begin{center}
\vspace*{-3ex}
\noindent\texttt{\footnotesize
BOS\sp f$_1$\sp f$_2$\sp and\sp f$_3$\sp \sp f$_4$\sp hand\sp it\sp f$_5$\sp over\sp f$_6$\sp to\sp f$_7$\sp f$_8$\sp you\sp f$_9$\sp EOS
}
\end{center}
where \texttt{f$_1$}, \texttt{f$_2$}, ... are encoder frames.
With such target sequence and interleaved speech/word-token embeddings, we can now train the model end-to-end with CE loss.


Fig.~\ref{fig:arch} contrasts how speech and word-token embeddings flow through the system for Speech ReaLLM compared to Speech LLM and RNN-T.
For ReaLLM (a), speech and text embeddings are sequentially interleaved, time index $t$ and labels index $k$ being independent variables.
For Speech LLM (b), the speech embeddings precede the text embeddings, i.e.~all speech must be
present before the first token can be emitted; hence Speech LLM is non-streaming.
RNN-T also has independent time and label indices $t$ and $k$, but while for ReaLLM we pursue exactly
one alignment (from the alignment teacher), RNN-T hypothesizes all valid alignments of label indices $k$ over time $t$.

\def\R{$\mathcal{R}$}
\def\L{$\mathcal{L}$}
\def\T{$\mathcal{T}$}
\def\S{$\mathcal{S}$}
\def\A{$\mathcal{A}$}
\def\P{$\mathcal{P}$}
\def\D{$\mathcal{D}$}

\section{Experiments and Results}
\subsection{Dataset}

We evaluated Speech ReaLLM using the well-known Librispeech benchmark \cite{librispeech}.
%
Librispeech consists of audio books in the public domain.
It includes a training set of 960h, as well as
an evaluation set (``test'') set and a development set (``dev''),
each consisting of an easier ``clean'' and a harder ``other'' subset, of 5.3 hours each.
%
No external language model was used.

\subsection{System parameters}

We compare Speech ReaLLM against two baseline architectures: (1) the non-streaming Speech LLM \cite{wu2023decoder,fathullah2023prompting}
and (2) an RNN-T \cite{graves2012} with joint network \cite{rnntwithjoiner}.
Unless otherwise noted, all models have about 80M parameters and are trained from scratch for up to 900 epochs, at max.~learning rate of 0.0005, with a tri-partite scheduler (warmup/holding/decay of 32/64/128 epochs).

All share the same 80-channel log-FBANK front-end with SpecAugment, and the same Streaming Conformer \cite{Shi2022-iu} encoder with 20 layers, a hidden dimension of 320, a widened FFN dimension of 2048, a temporal convolution of 7 frames, and a large Conformer segment size of 1.92 seconds with left and right context of 1 and 0.96 seconds, resp., with relative position embeddings \cite{shaw2018}.
The Conformer operates on 20-ms frames from two-way frame stacking; its
output is further frame-reduced by stacking and projection to 240 ms\footnote{240 ms is roughly the duration of one spoken word.} for Speech ReaLLM and Speech LLM, and to 60 ms for the RNN-T.


In Speech ReaLLM and Speech LLM, the encoder is followed by a Llama-2 decoder stack \cite{touvron2023llama2}. For training from scratch on 960h,
we reduced it to only two layers, an 8-head Transformer with dimension of 256, and a FFN dimension of 2048.
The RNN-T has a two-layer predictor LSTM of the same dimensions.
The output vocabulary is 4096 sentence pieces.

\subsection{Results}

Table \ref{tab:main} shows the main result.
Our 80M-parameter Speech Rea\-LLM has WERs of 3.0 and 7.4\% for test-clean and other, resp., which is within 9\% relative of the 3x larger LAS-SpecAugment model \cite{Park2019-tz}, a well-known baseline model without additional training data, external LM, auxiliary losses, or other tricks besides SpecAugment.\footnote{SoTA on Librispeech is much better, as low as 1.4\% (test-clean, no LM) with a non-streaming Conformer, wav2vec, and semi-supervised data \cite{Zhang2020-gk}; techniques not relevant to the investigation at hand.
}

\begin{table}[h]
\begin{center}
\footnotesize
\begin{tabular}{|c|l|c|c|c|c|c|c|}
\hline
\b Id\b& Architecture            &\b Params\b&\b Strea-\b& \multicolumn{2}{c}{\b WER test\b} & \multicolumn{2}{c|}{\b WER dev\b}  \\
   & family                      & [M]   &\b ming?\b &\b clean\b&\b other\b&\b clean\b&\b other\b\\ \hline\hline
\b\R\b& Speech ReaLLM                   & 81.6  & Yes   & 3.0   & 7.4   & 2.7   & 7.6  \\ \hline\hline 
\b\L\b& Speech LLM                    & 81.6  & No    & 4.8   & 8.0   & 4.2   & 8.3  \\ \hline       
\b\T\b& RNN-T                           & 79.3  & Yes   & 3.6   & 9.4   & 3.3   & 9.6  \\ \hline\hline 
\b\P\b& \R~$+$ pre-trained              & 7B    & Yes   & 4.7   & 9.1   & 4.2   & 9.5  \\ \hline\hline 
\b\S\b& LAS-SpecAug. \cite{Park2019-tz} &\b $>$270\b& No    & 2.8   & 6.8   & n/a   & n/a  \\ \hline
\end{tabular}
\end{center}
\vspace*{-0.5em}
\caption{Speech ReaLLM works. Comparing 80M Speech Rea\-LLM (\R, this paper) with non-streaming Speech LLM (\L), RNN-T (\T), and a matched public baseline (\S, LAS-SpecAugment); all trained from scratch on Librispeech only.
A 7B Speech ReaLLM with a pre-trained fine-tuned Llama-7B decoder (\P) also works in principle, but with some regression.}
\label{tab:main}
\vspace*{-2em}
\end{table}

A non-streaming 80M Speech LLM model \cite{wu2023decoder,fathullah2023prompting}, at 4.8 and 8.0\% is worse, but as we will see in section \ref{sec:length}, this is due to particular problems with the longest utterances, without which it would outperform Speech ReaLLM.
The RNN-T, at 3.6 and 9.4\%, performs worst of all models.
{\bf Speech ReaLLM is a viable new ASR architecture.}


\subsubsection{Can a ``decoder-only'' LM learn the flow of time?}
\label{sec:aligned}

In a simplified experiment, we trained a non-streaming {\em Time-aligned Speech LLM} model that, like Speech LLM, gets the entire speech in its prompt, but like Speech ReaLLM, has BLANK symbols in the target strings. Do decoded BLANK symbols represent time alignment?
Table \ref{tab:align} shows Alignment Error Rate (AER---how many words were expected vs.~decoded in each 240-ms speech chunk) and Length Error Rate (LER how many utterances had a wrong predicted length).
The alignments are good, with AERs below 7\%, and the LERs under 2\%.
%
{\bf A ``decoder-only'' LM is able to learn and reproduce the flow of time.}
We also see that Time-Aligned Speech LLM itself is a functional ASR model, with no WER regression.

\begin{table}[h]
\begin{center}
\footnotesize
\def\b{\hspace*{-1ex}}
\begin{tabular}{|c|l|c|c|c|c|c|c|c|}
\hline
\b Id\b& Architecture              & \multicolumn{2}{c|}{AER dev} & \multicolumn{2}{c|}{LER dev} & \multicolumn{2}{c|}{WER dev}  \\
   & family                        & \b clean\b&\b other\b     & \b clean\b&\b other\b     &\b clean\b&\b other\b\\ \hline\hline
\b\L\b& Speech LLM               & n/a   & n/a & n/a & n/a & 4.2   & 8.3  \\ \hline 
\b\A\b& \L~+ Time-aligned             & 4.9   & 6.5 & 1.0 & 1.4 & 4.1   & 7.6  \\ \hline 
\end{tabular}
\end{center}
\vspace*{-0.5em}
\caption{A Speech LLM architecture can indeed be trained to accurately reproduce the times at which words were spoken.}
\label{tab:align}
\vspace*{-2em}
\end{table}

\subsubsection{Can a pre-trained Llama-7B be fine-tuned to learn time?}

To validate ReaLLM's compatibility with real large pre-trained LLMs, we fine-tuned the ReaLLM mechanism into a frozen pre-trained 7B Llama-2 (\texttt{llama-2-7b-hf}), via rank-16 LoRA adapters on all Transformer projections.
The encoder is also frozen, from row \R, with a two-layer adapter MLP to map 256-dimensional embeddings to the Llama's 4096-dimensional space. We arbitrarily repurposed a character of the existing vocabulary, the underscore, as BLANK. A total of 27.3M trainable parameters are fine-tuned on Librispeech for 27 epochs.


Row \P~in Table \ref{tab:main} shows that it works, although the WERs of 4.2 to 9.5\% are in the upper range compared to 80M models that had been trained from scratch.
Inspecting the ASR output reveals that the model frequently inserts a random hallucinated word at the start of the utterance, while otherwise doing well after that.
We feel it is fair to conclude that, at least in principle,
{\bf a 7B Llama model can be fine-tuned to learn the ReaLLM behavior}, although further analysis and understanding is required.


\begin{table}[b]
\begin{center}
\footnotesize
\def\b{\hspace*{-1ex}}
\begin{tabular}{|c|l|c|c|c|c|}
\hline
\b Id\b& Architecture                  & Label  &\b Right\b& \multicolumn{2}{c|}{WER dev}  \\
   & family                            & delay  &\b context\b &\b clean\b&\b other\b\\ \hline\hline
\b\R\b& Speech ReaLLM                  & 0 ms   &\b 960 ms\b& 2.7   & 7.6  \\ \hline 
\b\D\b& \R~$+$ label delay $-$ right context &\b 480 ms\b&\b 480 ms\b& 4.4   & 9.1  \\ \hline 
\end{tabular}
\end{center}
\vspace*{-0.5em}
\caption{Shifting 480 ms of future context from the encoder to the decoder leads to accuracy regression.}
\label{tab:loss}
\vspace*{-1.5em}
\end{table}

\subsubsection{Time alignment and the loss function}

Is it better to provide the ASR system future context via the encoder or the decoder?
We tried predicting output words at a delay of 2 tokens (480 ms), giving the decoder access to two future labels; while reducing the Conformer right context by the same amount.
Table \ref{tab:loss} shows that this does not work as well, possibly because the decoder is much smaller. It may be different with a deeper decoder trained on more training data.

\subsubsection{Utterance length}
\label{sec:length}

Table \ref{tab:length} shows the dev-set results from Table \ref{tab:main} broken out by grouping the test utterances into three length groups.
Amongst the 80M models trained from scratch,
the streaming {\bf Speech Rea\-LLM (\R) performs consistently across lengths}, as does the RNN-T (\T).

\begin{table}[h]
\begin{center}
\footnotesize
\def\b{\hspace*{-1ex}}
\begin{tabular}{|c|l|c|c|c|c|c|c|}
\hline
\b Id\b& Architecture              & \multicolumn{2}{c|}{Short} & \multicolumn{2}{c|}{Medium} & \multicolumn{2}{c|}{\b Longest 100\b}  \\
       & family                    & \multicolumn{2}{c|}{WER dev} & \multicolumn{2}{c|}{WER dev} & \multicolumn{2}{c|}{WER dev}  \\
   &                               &\b clean\b&\b other\b& clean&\b other\b& clean&\b other\b\\ \hline\hline
\b\R\b& Speech ReaLLM              & 3.1   & 8.0 & 2.4   & 7.3 & 2.6   & 7.4  \\ \hline  
\b\L\b& Speech LLM               & 2.8   & 7.2 & 2.3   & 6.1&\b17.8&\b22.0 \\ \hline  
\b\A\b& \L~+ Time Aligned          & 2.7   & 6.6 & 2.2   & 6.2&\b17.3&\b17.4 \\ \hline  
\b\T\b& RNN-T                      & 4.5 &\b10.0 & 3.1   & 9.4 & 3.1   & 9.3  \\ \hline\hline  
\b\P\b& \R~$+$ pre-trained         & 4.9 &\b10.4 & 3.7   & 8.8 & 3.8   & 8.7  \\ \hline\hline  
    \multicolumn{2}{|r|}{Avg.~length\b}  &\b \em 12.6\b&\b \em 11.6\b& \em 32.0&\b \em 28.5\b& \em 61.75&\b \em 54.7\b\\ \hline
\end{tabular}
\end{center}
\vspace*{-0.5em}
\caption{The Speech LLM architecture stumbles over long sentences. Speech ReaLLM and RNN-T are unfazed.}
\label{tab:length}
\vspace*{-2em}
\end{table}

The non-streaming Speech LLM (\L), however, struggles, with WERs above 17\% on the longest 100 utterances. Its output has words dropped and in wrong order.
The Time-Aligned Speech LLM (\T, Sec.~\ref{sec:aligned}) exhibits almost the same pattern.
Additionally, it so happened that, by means of a configuration bug, we incidentally trained a Llama-ReaLLM ``chimera''
that received the speech frames in both the left context (like Llama) and interleaved
ReaLLM); this chimera exhibited a similar issue, except less pronounced.
This points to a challenge with modeling the long-span dependency into the prompt---the two-layer speech decoder may be too small to model this, or undertrained given the lack of long utterances in the training data. Both Speech ReaLLM and RNN-T gracefully avoid this issue, presumably by virtue of their different modes of operation.

We also find that a 7B
Speech ReaLLM with a pre-trained Llama-2 decoder handles long utterances just fine.

%



\subsubsection{Inference cost, Real-time Factor, and Beam Search}

\def\E{\mathcal{E}}
\def\D{\mathcal{D}(T)}
\def\O{\mathcal{O}}
\def\f{\mathcal{O}}

How does the inference cost of Speech ReaLLM compare to Speech LLM and RNN-T?
A back-of-the-envelope estimate of inference cost---for greedy search to keep it simple---is:
\begin{eqnarray*}
C_\mathrm{common} &=& T \cdot \E + U \cdot \D + U \cdot \O \\
C_\mathrm{ReaLLM} &=& C_\mathrm{common} + T f \cdot \D + T f \cdot \O \\
C_\mathrm{Llama}  &=& C_\mathrm{common} + T f \cdot \D \\
C_\mathrm{RNN-T}  &=& C_\mathrm{common} + T f' \cdot \O
\end{eqnarray*}
with duration $T$
; encoder frame rate $f$ for LLM and $f'$ for RNN-T; number of output tokens $U$; encoder cost per second $\E$; average decoder/predictor cost per token $\D$ (length-dependent for Transformers), and output/joint-network cost $\O$.

The expressions are confirmed by Table \ref{tab:rtf}, which shows real-time factors measured on a typical x64 development server (Intel Core/Broadwell, 16 MB cache, 1.995 GHz, CentOS 9) for single-threaded fp32 decoding in interpreted Python via PyTorch eager mode,
on 11 equidistantly selected utterances from the length-sorted dev-other set, on average 7.4s long.

At an RTF of 0.94, Speech ReaLLM is a bit more costly than Speech LLM at 0.81, because it runs the output layer also for each input chunk. Furthermore, Speech LLM pre-fills all speech embeddings at once, benefitting from weight reuse.
RNN-T in our case is slower, 1.0, than Speech ReaLLM, as it requires a higher encoder frame rate $f'\,=\,4 f$ for good accuracy.
%
Overall, the differences are limited as the runtime is dominated by the 20-layer encoder.

\begin{table}[t]
\begin{center}
\footnotesize
\def\b{\hspace*{-1ex}}
\begin{tabular}{|c|l|c|c|c|c|c|c|}
\hline
\b Id\b& Architecture              & \multicolumn{3}{c|}{Beam Search (4)} & \multicolumn{3}{c|}{Greedy Search}  \\
   & family                        & RTF   & \multicolumn{2}{c|}{WER dev}& RTF & \multicolumn{2}{c|}{WER dev}\\
   &                               &       &\b clean\b&\b other\b&     &\b clean\b&\b other\b\\ \hline\hline
\b\R\b& Speech ReaLLM              & 0.94   & 2.7   & 7.6 & 0.79  & 2.8   & 7.8  \\ \hline\hline  
\b\L\b& Speech LLM               & 0.81   & 4.2   & 8.3 & 0.73   & 4.5   & 7.9  \\ \hline  
\b\T\b& RNN-T                      & 1.00   & 3.3   & 9.6 & 0.90   & 3.5   & 10.0  \\ \hline  
\end{tabular}
\end{center}
\vspace*{-0.5em}
\caption{Real-time factor and sensitivity of Speech ReaLLM and the Speech LLM and RNN-T baselines to search beam.}
\label{tab:rtf}
\vspace*{-2.5em}
\end{table}

{\bf 80M Speech ReaLLM runs in real time} on a dev server.
With its two narrow decoder layers, it is approximately as large as an RNN-T model one could run on a wearable device. Thus, we estimate that an 8-bit quantized 80M Speech ReaLLM could comfortably run in real time on a wearable processor with multi-threaded or hardware-accelerated execution.

Inference can be sped up by using greedy search instead of beam search.
Table \ref{tab:rtf} shows a slight accuracy regression of 3--10\% relative from greedy search for all models, but the runtime benefits are drowned out by the dominating speech encoder.

\section{Conclusions}

We have introduced a new way of using ``decoder-only'' models (``LLMs'') to imbue them with the ability to process inputs in a real-time streaming fashion, via an RNN-T-like BLANK mechanism.
We have validated this approach, which we termed {\em Rea\-LLM} for ``real-time LLM'',
for real-time streaming ASR.

{\em Speech ReaLLM} has been found to be a viable new ASR architecture.
Its 80M variant runs in real time and achieves accuracy on par with or better than a non-streaming baseline Speech LLM architecture and an RNN-T of the same size,
and close to a 3x larger non-streaming LAS-SpecAugment AED baseline.
Unlike Speech LLM and AED architectures, it is designed to handle streaming input without explicit end-pointing,
and indeed generalizes well to long utterances which are poorly represented in the training data.
Unlike RNN-T, it has a simple loss function.
We also show that the ReaLLM architecture can learn to represent and reproduce the passing of time.

We also find that the ReaLLM mechanism can be fine-tuned into a 7B Llama-2 decoder,
although the resulting system is not as good as the small models trained from scratch.
More research is needed into making best use of the pre-trained 7B LLM. For example, can one replace the distinct BLANK symbol by EOS? Can such a model benefit from textual context, such as an audio stream's metadata or other context? And to what degree does the fine-tuned LLM retain its original capabilities?

We close with the expectation that the ReaLLM architecture has potential beyond ASR and may open up many interesting applications beyond speech transcription, from more natural human-computer dialogs to more general real-time aware intelligent assistants---possibly taking us one baby step closer towards AGI?

\section{Acknowledgments}
The authors would like to thank Egor Lakomkin and Zeeshan Ahmed for helping with the LoRA experiments.

\clearpage

\bibliographystyle{IEEEtran}
\bibliography{mybib}

\end{document}